\documentclass[runningheads]{llncs}
\usepackage{graphicx}

\usepackage{tikz}
\usepackage{comment}
\usepackage{amsmath,amssymb} 
\usepackage{color}

\usepackage{amsmath,amssymb}
\usepackage{epsfig}
\usepackage{subfigure}
\usepackage{epstopdf}
\usepackage{chngpage}
\usepackage{array}
\usepackage{makecell}
\usepackage{multirow}
\usepackage[noend]{algpseudocode}
\usepackage{algorithmicx}
\usepackage{algorithm}
\usepackage{url}
\usepackage{booktabs}
\usepackage{caption}
\usepackage{ragged2e}
\usepackage{hyperref}
\usepackage[misc]{ifsym}

\begin{document}
\pagestyle{headings}
\mainmatter
\def\ECCVSubNumber{100}  

\title{GeoLayout: Geometry Driven Room Layout Estimation Based on Depth Maps of Planes} 

\titlerunning{GeoLayout: Geometry Driven Room Layout Estimation}
%
\author{Weidong Zhang\inst{1,2}\and
Wei Zhang\inst{1}\thanks{Corresponding author. E-mail: davidzhang@sdu.edu.cn}\and
Yinda Zhang\inst{3}}

\authorrunning{W. Zhang et al.}
%

\institute{School of Control Science and Engineering, Shandong University \and
School of Communications and Information Engineering, Xi'an University of Posts \& Telecommunications \and
Google Research\\}

\maketitle

\begin{abstract}
The task of room layout estimation is to locate the wall-floor, wall-ceiling, and wall-wall boundaries. Most recent methods solve this problem based on edge/keypoint detection or semantic segmentation. However, these approaches have shown limited attention on the geometry of the dominant planes and the intersection between them, which has significant impact on room layout. In this work, we propose to incorporate geometric reasoning to deep learning for layout estimation. Our approach learns to infer the depth maps of the dominant planes in the scene by predicting the pixel-level surface parameters, and the layout can be generated by the intersection of the depth maps. Moreover, we present a new dataset with pixel-level depth annotation of dominant planes. It is larger than the existing datasets and contains both cuboid and non-cuboid rooms. Experimental results show that our approach produces considerable performance gains on both 2D and 3D datasets.

\end{abstract}

\section{Introduction}

An indoor scene differs from the natural scenes in that it usually contains dominant planes such as floor, ceiling and walls. These planes are likely to be orthogonal to each other. Hence the spatial structure of an indoor scene tends to show some regularity and can be represented by the room layout. Currently, the task of room layout estimation is to locate the wall-floor, wall-ceiling, and wall-wall boundaries. It can provide a useful prior for a wide range of computer vision tasks, such as scene reconstruction~\cite{martin20143d,camplani2013depth,izadinia2017im2cad} and augmented reality~\cite{xiao2014reconstructing,karsch2011rendering,liu2015rent3d}.

Recent methods achieve significant performance gains, which primarily focus on learning the feature maps with deep networks like fully convolutional networks (FCNs)~\cite{long2015fully1}. One popular idea is to learn the wall-floor, wall-ceiling, and wall-wall edges~\cite{mallya2015learning,ren2016coarse,zhao2017physics}. Another is to learn the semantic surface labels such as floor, ceiling, front wall, left wall, and right wall~\cite{dasgupta2016delay,zhang2019edge}. Besides, there are also methods trying to infer the layout corners (keypoints)~\cite{lee2017roomnet,zou2018layoutnet}. However, the gathered bottom-up information from edge/keypoint detection or semantic segmentation may not reflect the underlying geometry of room layout, e.g., orthogonal planes.

\begin{figure}
\begin{center}
\includegraphics[width = 1.0\columnwidth]{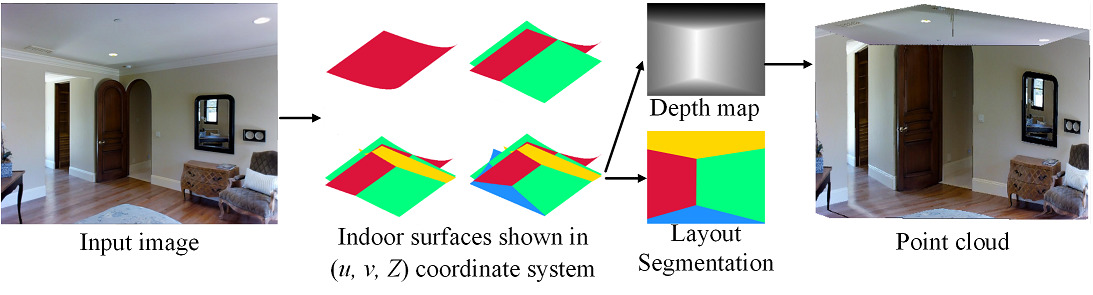}
\end{center}
\caption{Room layout estimation based on depth maps of the planar surfaces.}
\label{fig:overview}
\end{figure}

Essentially, the desired boundary between two surfaces appears because the two planes in 3D space intersect in a line. This motivates us to focus on the geometric model of the dominant surfaces (e.g., the floor, ceiling and wall) in the indoor scene. With this key insight, we propose to predict the depth maps of the dominant surfaces, and generate the layout by the intersection of the depth maps, as shown in Fig.~\ref{fig:overview}. We first analyse the projection principle of a 3D plane into the depth map and obtain the representation without explicit camera intrinsics to parameterize the depth of a plane. Compared to the general 3D coordinate systems (e.g., the camera coordinate system), our parameterization can omit the need for the camera intrinsic parameters. It also makes the method applicable to the existing layout datasets whose intrinsic parameters are not provided like Hedau~\cite{hedau2009recovering} and LSUN~\cite{yinda2016lsun}. Then we train a deep network to predict the pixel-level surface parameters for each planar surface. The pixel-level parameters are further aggregated into an instance-level parameter to calculate the corresponding depth map, and the layout can be generated based on the predicted depth maps. Our method generally requires the depth map of the planar surfaces for learning. However, with our parameterization and geometric constraints, the model can also be trained with only 2D segmentation.

However, the existing datasets for layout estimation do not fully support the learning of the proposed 3D geometry-aware model as the 3D labels are not provided. Besides, all the images are exhibiting simple cuboid layout only. These shortcomings of the datasets severely limit the development of layout estimation algorithms and the practical applications. Therefore, we produce a new dataset for room layout estimation providing pixel-level depth annotation of the dominant planar surfaces. The ground truth is gathered semi-automatically with a combination of human annotation and plane fitting algorithm, and the dataset contains indoor scenes with complex non-cuboid layout.

The major contributions of this work are summarized as follows: (1) We propose to incorporate geometric reasoning to deep learning for the task of layout estimation, which is reformulated as predicting the depth maps of the dominant planes. (2) We demonstrate the proposed model can be effectively trained to predict the surface parameters, and it can also improve 2D layout performance with the learned 3D knowledge. (3) A dataset with 3D label for layout estimation is presented. The dataset is in large scale, complementary to previous datasets, and beneficial to the field of room layout estimation.

\section{Related Work}

The current work on layout estimation can be divided into two types according to whether the 3D spatial rules are exploited.

\noindent\textbf{2D based layout estimation.} The layout estimation problem was first introduced by Hedau et al.~\cite{hedau2009recovering}, which consisted of two stages. First, a series of layout hypotheses were generated by ray sampling from the detected vanishing points. Second, a regressor was learned to rank the hypotheses. Later on, some methods were proposed to improve the framework~\cite{ramalingam2013manhattan,wang2013discriminative,pero2012bayesian,del2013understanding}, such as using different hand-crafted features and improving the hypotheses generation process.

Recently, the CNN and FCN-based methods were proposed for layout estimation and showed dramatic performance improvements on benchmark datasets. Mallya and Lazebnik~\cite{mallya2015learning} trained a FCN~\cite{long2015fully1} to predict the edge maps, which are used for layout hypotheses generation and ranking. Ren et al.~\cite{ren2016coarse} used the edge map as a reference for generating layout hypotheses based on the vanishing lines, undetected lines, and occluded lines. Dasgupta et al.~\cite{dasgupta2016delay} instead predicted the pixel-level semantic labels with FCN, and the layout estimates are further optimized by vanishing lines. Zhang et al.~\cite{zhang2019edge} jointly learned the edge maps and semantic labels using an encoder-decoder network. The layout hypotheses were then generated and optimized based on the two information. Zhao et al.~\cite{zhao2017physics} transferred the semantic segmentation model to produce edge features and proposed a physics inspired optimization inference scheme. Lee et al.~\cite{lee2017roomnet} adopted an end-to-end framework for layout estimation by predicting the locations of the layout keypoints. These methods generally predict the layout in 2D space, and the 3D knowledge of indoor scenes are usually ignored.

\noindent\textbf{3D based layout estimation and related work.} Lee et al.~\cite{lee2009geometric} proposed the "Indoor World" model based on the Manhattan world assumption and symmetric floor and ceiling assumption. The model represented the scene layout in 2D space and could be translated into a 3D model by geometric reasoning on the configuration of edges. Choi et al.~\cite{choi2013understanding} trained several graph models that fused the room layout, scene type and the objects in 3D space. Zhao and Zhu~\cite{zhao2013scene} applied a hierarchical model that characterized a joint distribution of the layout and indoor objects. The layout hypotheses were evaluated by the 3D size and localization of the indoor objects. Guo et al.~\cite{guo2015predicting} trained five SVMs for the five layout categories using appearance, depth, and location features. These methods have exploited 3D spatial rules or considered the 3D relationship between the room layout and indoor objects, but none of these work has focused on the 3D geometry of the planar surfaces.

The 3D plane detection and reconstruction problem aims to segment plane instances and recover 3D plane parameters from an image, which is somewhat similar to layout estimation. The methods can be divided into two groups. The geometry-based methods~\cite{delage2007automatic,barinova2008fast,micusk2008towards} extract geometric cues such as vanishing points and line segments to recover 3D information. The appearance-based methods~\cite{fouhey2014unfolding,haines2014recognising,liu2018planenet,liu2019planercnn,yang2018recovering,yu2019single} infer the 3D information based on the appearance and do not rely on the assumptions about the scenes. Specifically, Liu et al.~\cite{liu2018planenet} proposed a deep network that simultaneously learned a constant number of plane parameters and instance-level segmentation masks. Yang and Zhou~\cite{yang2018recovering} reformulated the problem and proved that the 3D plane fitting losses could be converted to depth prediction losses, and therefore did not require ground truth 3D planes. Yu et al.~\cite{yu2019single} presented a proposal-free instance segmentation approach that first learned pixel-level plane embeddings and then applied the mean shift clustering to generate plane instances. Since these methods are purely 3D based and require the camera intrinsic parameters to work, they cannot be applied to the current layout datasets like Hedau and LSUN.

\section{Method}

\begin{figure*}
\begin{center}
\includegraphics[width = 1.0\textwidth]{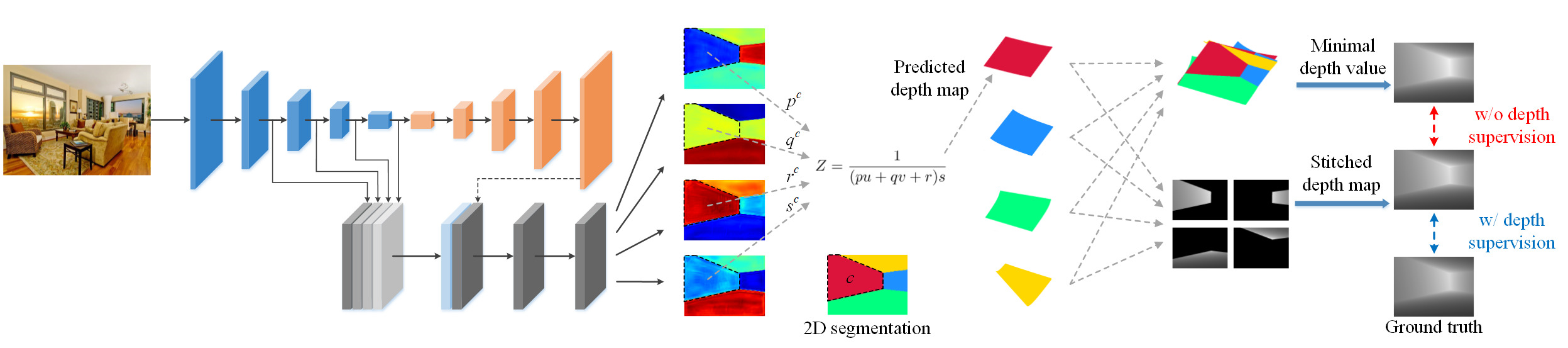}
\end{center}
\caption{An illustration of our method that can be trained w/wo depth supervision. Given an input image, the pixel-level surface parameters are predicted by the network. They are aggregated into several instance-level parameters to produce the depth maps of the planar surfaces. Based on the 2D segmentation, these depth maps can be combined into a stitched depth map, which is evaluated by either the ground truth (w/ depth supervision) or the minimal depth value of the predicted depth maps for each pixel localization (w/o depth supervision).}
\label{fig:process}
\end{figure*}

In this work, we intend to solve the layout estimation problem by predicting the depth maps of the dominant planes (e.g., floor, walls, ceiling) in a room. Then the layout can be obtained by the intersection of the depth maps of planar surfaces that intersect each other. In Section~\ref{sec:param}, we first analyze the depth map of a plane and give the general equation in the $(u,v,Z)$ coordinate system, which can be used to parameterize the depth map of an arbitrary plane. Then we use a deep network to learn the surface parameters of the dominant planes and generate layout estimates. The illustration of our method is shown in Fig.~\ref{fig:process}.

\subsection{Parameterizing Depth Maps of Planes}
\label{sec:param}

A 3D plane in the camera coordinate system can be represented with the equation: $aX+bY+cZ+d=0$, where $(a,b,c)$ is the normal vector and $d$ is the distance to the camera center. A 3D point can be projected onto the image plane via perspective projection, i.e., $u=f_x\frac{X}{Z}+u_0,\quad v=f_y\frac{Y}{Z}+v_0$, where $u$, $v$ are the pixel coordinates and $f_x$, $f_y$, $u_0$, $v_0$ are the camera intrinsic parameters, with $f_x$ and $f_y$ the focal lengths and $u_0$ and $v_0$ the coordinates of the principal point. Based on the perspective projection, The planar equation can be rewritten as: $\frac{1}{Z}=-\frac{a}{f_xd}u - \frac{b}{f_yd}v+ \frac{1}{d}(\frac{a}{f_x}u_0+\frac{b}{f_y}v_0-c)$. Apparently, the inverse depth value $\frac{1}{Z}$ is proportional to the pixel coordinates $u$ and $v$ in the depth map. With the above observation, the depth map of a plane can be parameterized without explicit camera intrinsics by three new parameters $\hat{p}$, $\hat{q}$, $\hat{r}$ as shown in Eq.~(\ref{eq:curve}).

\begin{equation}
\begin{aligned}
\label{eq:curve}
Z=\frac{1}{\hat{p}u+\hat{q}v+\hat{r}}.
\end{aligned}
\end{equation}

In practice, the global scale of an indoor scene is ambiguous, which makes the three parameters involved with the scale. Hence, we introduce a scale factor $s=\sqrt{\hat{p}^2+\hat{q}^2+\hat{r}^2}$ and apply normalization to $\hat{p}$, $\hat{q}$, $\hat{r}$, i.e., $p=\hat{p}/s$, $q=\hat{q}/s$, $r=\hat{r}/s$. The normalized parameters $p$, $q$, $r$ are therefore scale-invariant. Finally, the modified equation is given as below:

\begin{equation}
\begin{aligned}
\label{eq:normalization}
Z=\frac{1}{(pu+qv+r)s}.
\end{aligned}
\end{equation}

\subsection{Learning Depth Maps of Planes}
\label{sec:3d}

We first introduce our method that learns the depth maps of planes with depth supervision in this section.

\noindent\textbf{Pixel-level surface parameter estimation.} As shown in Eq.~(\ref{eq:normalization}), the depth map of a plane in the real world can be parameterized using $p$, $q$, $r$ and $s$. Motivated by this, we train a deep network to predict the pixel-level surface parameters of the input image. We implement the surface parameter estimation network based on~\cite{hu2019revisiting}, which is originally designed for monocular depth estimation. The network consists of four modules: an encoder, a decoder, a multi-scale feature fusion module, and a refinement module. We replace the last layer of the refinement module by four output channels. For an input color image, the network outputs four heat maps representing the pixel-level surface parameters. It is worth noting that the network can be replaced to any architecture for pixel-wise prediction such as PSPNet~\cite{zhao2017pyramid} and FCN~\cite{long2015fully1}.

With the ground truth depth map of the dominant planes $Z^*_i$, we transform Eq.~(\ref{eq:curve}) to calculate the target parameters $\hat p_i^*$, $\hat q_i^*$, $\hat r_i^*$ at the $i_{th}$ pixel:

\begin{equation}
\begin{aligned}
\label{eq:inv}
\hat{p}_{i}^{*}&= \triangledown _u  (1/Z^*_i), \\
\hat{q}_{i}^{*}&= \triangledown _v  (1/Z^*_i), \\
\hat r_i^*&=1/Z^*_i-\hat p_i^*u_i-\hat q_i^*v_i,
\end{aligned}
\end{equation}
where $\triangledown _u (1/Z^*_i)$ represents the spatial derivative of $(1/Z^*_i)$ w.r.t. $u$ computed at the $i_{th}$ pixel, and so on. Following the same normalization operation, the scale factor $s_i^*$ is calculated and the normalized parameters $p_i^*$, $q_i^*$, $r_i^*$ are obtained. Let $P_i=(p_i, q_i, r_i, s_i)$ be the predicted parameters at the $i_{th}$ pixel and $P_i^*$ denotes the ground truth. We use L1 loss to supervise the regressed parameters for each pixel:

\begin{equation}
\begin{aligned}
\label{eq:3dl1}
\mathcal{L}_p = \frac{1}{n}\sum_{i=1}^{n}\left \|  P_i- P_i^* \right \|.
\end{aligned}
\end{equation}

Besides, the surface parameters belonging to the same surface should be close together, while the parameters from different surfaces should be far apart. To this end, we employ the discriminative loss proposed in~\cite{de2017semantic}. The loss function includes two terms: a variance term to penalize the parameter that is far from its corresponding instance center, and a distance term to penalize the pairs of different instance centers that are close to each other.

\begin{equation}
\begin{aligned}
\label{eq:discrim}
\mathcal{L}_d = \mathcal{L}_{var} + \mathcal{L}_{dist},
\end{aligned}
\end{equation}

\begin{equation}
\begin{aligned}
\label{eq:var}
\mathcal{L}_{var}=\frac{1}{C}\sum_{c=1}^{C}\frac{1}{n_c}\sum_{i=1}^{n_c}\max(\left \|  P_i- P^c \right \|-\delta_v,0),
\end{aligned}
\end{equation}

\begin{equation}
\begin{aligned}
\label{eq:dist}
\mathcal{L}_{dist}= \frac{1}{C(C-1)}\mathop{\sum_{c_A=1}^{C}\sum_{c_B=1}^{C}}_{c_A\neq c_B}\max(\delta_d-\left \| P^{c_A}- P^{c_B} \right \|,0),
\end{aligned}
\end{equation}
where $C$ is the number of planar surfaces in the ground truth, $n_c$ is the number of pixels in surface $c$, $P^c$ is the mean of the pixel-level parameters belonging to $c$, $\delta_v$ and $\delta_d$ are the margins for the variance and distance loss, respectively. Here we employ the variance loss $\mathcal{L}_{var}$ to encourage the estimated parameters to be close within each surface. At last, we extract $P^c$ as the instance-level parameters to generate the depth map of surface $c$.

\noindent\textbf{Depth map generation.} We found that exploiting the ground truth to supervise the predicted depth map makes the training more effective. For surface $c$ with the predicted instance-level parameters $P^c=(p^c, q^c, r^c, s^c)$, its corresponding depth map $Z^c$ can be produced using Eq.~(\ref{eq:normalization}), i.e., $Z^c_i=1/[(p^c u_i + q^c v_i + r^c)s^c]$. In the training stage, the ground truth 2D segmentation $l^*$ is used to combine the predicted depth maps into a stitched depth map, which is evaluated by the ground truth as below:

\begin{equation}
\begin{aligned}
\label{eq:loss_layout}
\mathcal{L}_z = \frac{1}{n}\sum_{i=1}^{n}\left \| \frac{1}{Z^{l^*_i}_i} - \frac{1}{Z^*_i} \right \|,
\end{aligned}
\end{equation}
where $Z^{l^*_i}_i$ is the $i_{th}$ pixel of the stitched depth map and $Z^*$ is the ground truth. We use the inverse depth in the loss function as it is linear w.r.t. the pixel coordinates, which makes the training more stable and smooth. Finally, the overall objective is defined as follows:

\begin{equation}
\begin{aligned}
\label{eq:loss_3d}
\mathcal{L}_{3D} = \mathcal{L}_p + \alpha \mathcal{L}_{var} + \beta \mathcal{L}_z.
\end{aligned}
\end{equation}

\subsection{Training on 2D Layout Datasets}
\label{sec:2d}

The current benchmark datasets Hedau~\cite{hedau2009recovering} and LSUN~\cite{yinda2016lsun} both use the ground truth 2D segmentation to represent the layout, and neither of them has ground truth depth. The supervised learning method in Section~\ref{sec:3d} is inapplicable for these datasets. In this section, we present a learning strategy that enables the model to be trained with only 2D segmentation.

\begin{figure}
\begin{center}
\subfigure[Cuboid layout]{
\includegraphics[width = 0.47\columnwidth]{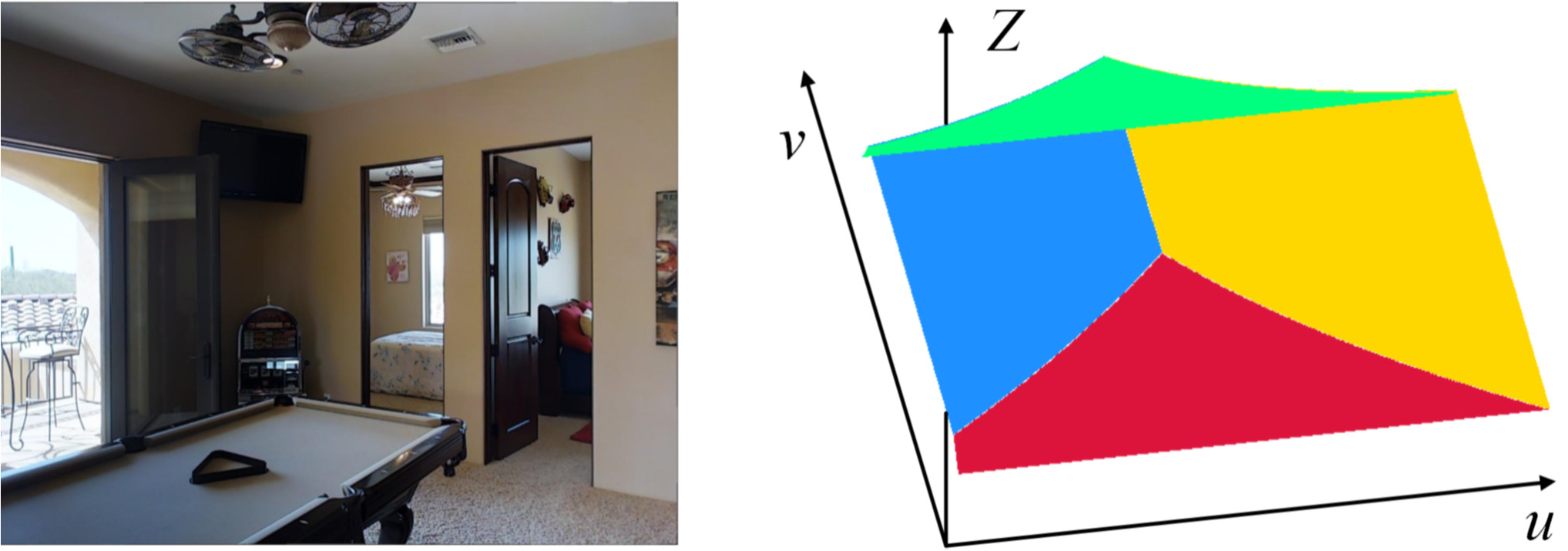}
}
\subfigure[Non-cuboid layout]{
\includegraphics[width = 0.47\columnwidth]{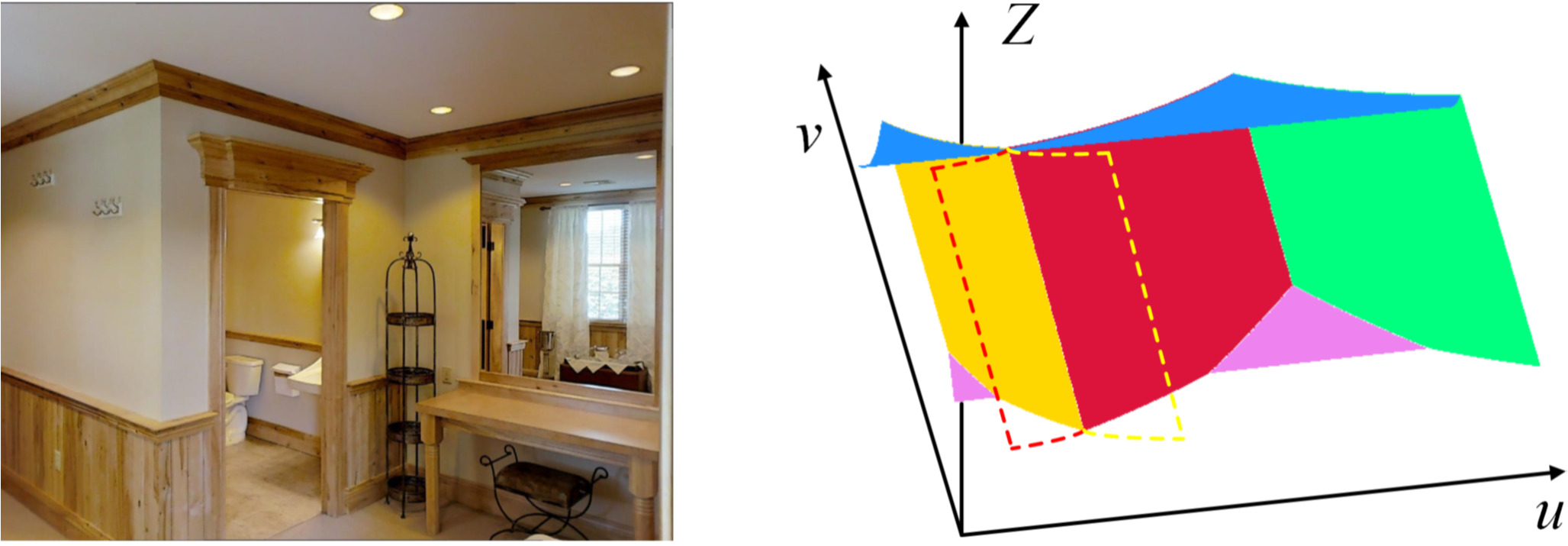}
}
\end{center}
\caption{Depth representation of the dominant planes in the $(u,v,Z)$ coordinate system. The layout of a cuboid room is determined by the nearest planar regions, but it is inapplicable for the non-cuboid room (see the red and yellow surfaces). }
\label{fig:cu}
\end{figure}

First, we employ the same network structure as in Section~\ref{sec:3d} for surface parameter estimation. We use the full discriminative loss $\mathcal{L}_d$ to constrain the predicted surface parameters. Here we assume that the indoor image has cuboid layout, which means the room can be represented as a box. Such assumption is tenable for Hedau and LSUN datasets and has been widely adopted by many previous work. Based on this assumption, an important observation is that the layout is determined by the nearest planar regions in the indoor scene. As shown in Fig.~\ref{fig:cu} (a), if representing the depth maps of all the dominant planes in the $(u,v,Z)$ coordinate system, each surface will have a depth value at each $(u, v)$ coordinate and the minimal value at each pixel will form the depth map that represents the layout. Also, the 2D segmentation map can be obtained according to which surface has the minimal depth value at each pixel. It is worth noting that the calculated depth map from Eq.~(\ref{eq:normalization}) may have negative values, which should be excluded. We simply switch to inverse depth and extract the maximum of $\frac{1}{Z^c}$ at each pixel to produce the layout segmentation and corresponding depth map:

\begin{equation}
\begin{aligned}
\label{eq:gen_layout}
\frac{1}{Z_i}=\max_c \frac{1}{Z^c_i}, \quad l_i=\arg\max_c \frac{1}{Z^c_i}.
\end{aligned}
\end{equation}

Here, $l_i$ is the generated pixel-level layout segmentation and $\frac{1}{Z_i}$ is the corresponding depth map. Since $\arg\max$ is not differentiable, it is unable to evaluate the generated layout estimates by the pixel error between $l_i$ and ground truth $l^*_i$. Instead, we encourage the stitched depth map to be consistent with the minimal depth value. The loss function for the predicted depth map is defined as below:

\begin{equation}
\begin{aligned}
\label{eq:loss_layout}
\mathcal{L}_z = \frac{1}{n}\sum_{i=1}^{n}\left \| \frac{1}{Z^{l^*_i}_i} - \frac{1}{Z_i} \right \|.
\end{aligned}
\end{equation}

However, we find that under the current objective the learned model tends to produce similar depth estimates for all the surfaces to reduce the loss. To deal with this problem, we propose a ``stretch" loss to increase the mutual distance between the depth maps as follows:

\begin{equation}
\begin{aligned}
\label{eq:loss_stretch}
\mathcal{L}_s = - \frac{1}{n}\sum_{i=1}^{n}\frac{e^{k /Z^{l^*_i}_i}}{\sum_{c=1}^{C}e^{k / Z^c_i}},
\end{aligned}
\end{equation}
where $k$ is a scale factor in the softmax operation. The ``stretch" loss encourages $1/Z^{l^*_i}_i$ to be much larger than the rest inverse depth values at $i_{th}$ pixel, and therefore similar depth estimates will be punished. The overall objective is defined as follows:

\begin{equation}
\begin{aligned}
\label{eq:loss_2d}
\mathcal{L}_{2D} = \mathcal{L}_d + \eta \mathcal{L}_z + \theta \mathcal{L}_s.
\end{aligned}
\end{equation}

It is worth noting that such learning strategy is inapplicable for the non-cuboid room, as shown in Fig.~\ref{fig:cu} (b). Besides, the generated depth map can only infer the relative depth information, yet the precise depth value is unavailable.

\subsection{Generating Layout Estimates}
\label{sec:post}

When the training stage is complete, a post-process step is employed to obtain the parameterized layout estimation results. Because of the discriminative loss (Eq.~(\ref{eq:discrim})-(\ref{eq:dist})), the predicted pixel-level surface parameters are likely to be piece-wise constant and can be easily grouped to produce a segmentation map representing the surface instances. We use standard mean-shift clustering as the number of clusters does not need to be pre-defined. After clustering, the small clusters with fewer than 1\% of the overall pixels are abandoned. Next we extract the mean of the parameters within each cluster to obtain the instance-level parameters. Then the depth map for each planar surface can be generated.

To find the true layout among the depth maps that intersect each other, we evaluate the layout estimates based on its consistency with the clustered segmentation. Specifically, we sort the depth maps of different surfaces according to ascending order for each pixel, while the index indicating the surface instance will constitute multiple layers of segmentation maps. Starting from the first layer, we compare the current segmentation with the clustered segmentation. For each region of the current segmentation, if the label is inconsistent with the dominant label of the clustered segmentation, we use the labels from the next layer to replace the inconsistent label. This process continues until the current segmentation is consistent with the clustered segmentation. Then the predicted layout segmentation, depth map, and the corresponding surface parameters are all available. With the intrinsic camera parameters, the 3D point cloud representing the layout can also be generated based on the depth map. Finally, the layout corners can be computed based on the equations of the predicted depth maps, i.e., a layout corner is the point of intersection among three surfaces, or two surfaces and an image boundary.

\section{Matterport3D-Layout Dataset}
\label{sec:dataset}

In this section, we introduce our large scale dataset with 3D layout ground truth for our training purpose, named Matterport3D-Layout. We use images from Matterport3D dataset~\cite{Matterport3D} as the dataset contains real photos from complex scenes, which provides good layout diversity. It also provides depth image that can be used to recover 3D layout ground truth. We annotate the visible region of each plane and use Eq.~(\ref{eq:curve}) for parameter fitting in each surface. Then the depth maps of planar surfaces can be calculated using Eq.~(\ref{eq:curve}).

\noindent\textbf{Annotation.} We first filter out the images without recognizable layout. Then we draw 2D polygons using LabelMe~\cite{russell2008labelme} on the visible regions of the floor, ceiling and walls for each image. The polygons on different surfaces have different semantic categories. We also abandon the images with surfaces completely occluded by the indoor objects as the true depth of the surfaces are unavailable.

\noindent\textbf{Layout generation.} Given the depth map and region annotation, we extract the depth value and pixel coordinates in each annotated region and employ RANSAC algorithm~\cite{fischler1981random} for the curved surface fitting to obtain the instance-level surface parameters. Then the layout can be generated in a similar way as described in Section~\ref{sec:post}.

The original Matterport3D dataset includes 90 different buildings, so we randomly split the dataset into training, validation and testing set according to the building ID. The training set includes 64 buildings with a total of 4939 images. The validation set includes 6 buildings with 456 images. The testing set includes the remaining 20 buildings with a total of 1965 images. All images have the resolution of $1024\times1280$. The dataset contains the following fields: (1) Color image; (2) Depth map of the planar surfaces; (3) 2D segmentation of layout; (4) Original depth map containing indoor objects; (5) Visible region annotation; (6) Intrinsic matrix of the camera; (7) Surface parameters for each plane $p$, $q$, $r$; (8) The coordinates of the layout corners $(u,v,Z)$; (9) Original surface normal.

\begin{figure}
\begin{center}
\includegraphics[width = 1.0\columnwidth]{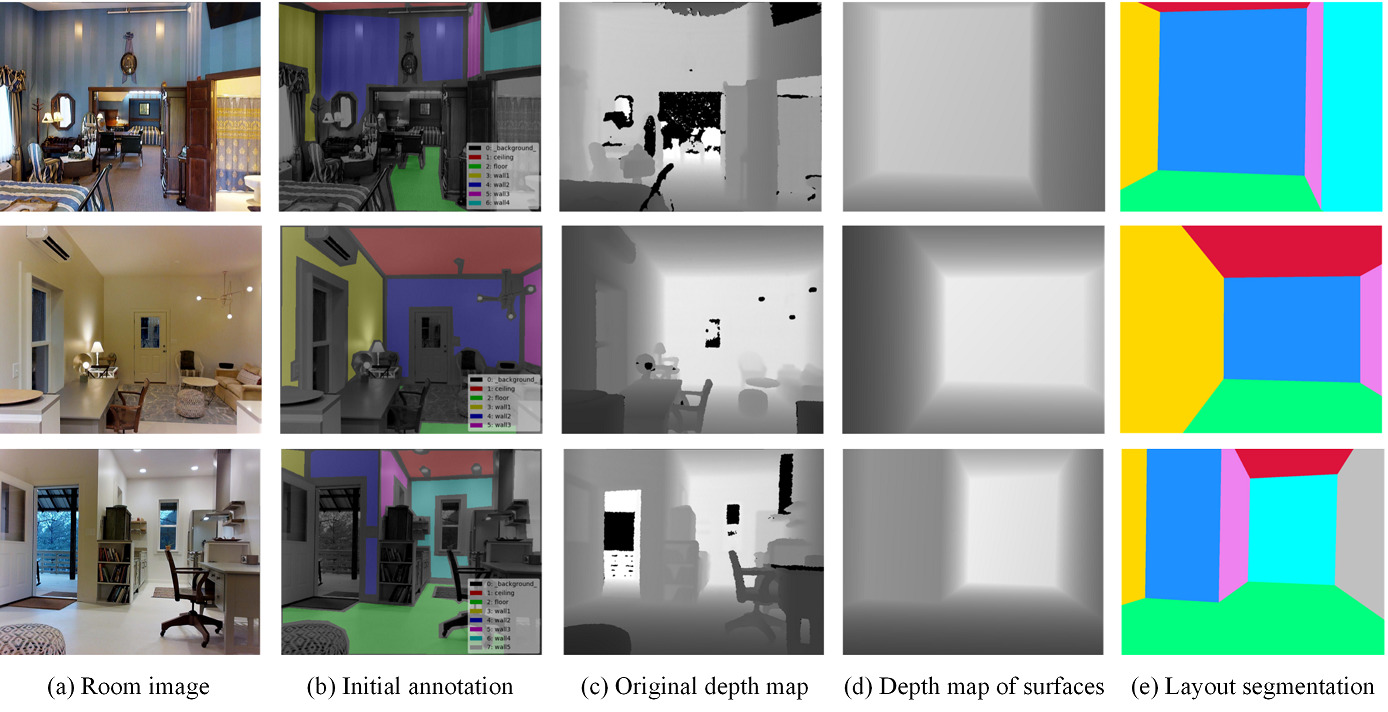}
\end{center}
\caption{Our Matterport3D-Layout dataset provides pixel-level depth label for the dominant planes.}
\label{fig:dataset}
\end{figure}

Fig.~\ref{fig:dataset} shows some examples of our dataset. Prior to our dataset, there are two benchmark layout datasets: Hedau~\cite{hedau2009recovering} and LSUN~\cite{yinda2016lsun}. Statistics of the existing datasets are summarized in Table~\ref{table:dataset_cmp}. As can be seen, the proposed dataset is the largest one and provides the richest kinds of ground truths. Besides, the proposed dataset contains non-cuboid layout samples which are absent in the other datasets. We hope this dataset can benefit the community and motivate the research about indoor layout estimation and related tasks.

\begin{table}
\begin{center}
\caption{A brief summary of existing datasets in layout estimation.}
\label{table:dataset_cmp}
\setlength{\tabcolsep}{1.2mm}{
\begin{tabular}{lccccc}
\hline
Dataset & Train & Val. & Test & Label & Layout type  \\
\hline
Hedau~\cite{hedau2009recovering} & 209 & - & 105 & seg. & cuboid \\
LSUN~\cite{yinda2016lsun} & 4000 & 394 & 1000 & \makecell[c]{seg. \& corner} & cuboid \\
Matterport3D-Layout & 4939 & 456 & 1965 & \makecell[c]{seg. \& corner \& depth} & cuboid \& non-cuboid \\
\hline
\end{tabular}}
\end{center}
\end{table}

\section{Experimental Results}

In this section, we first evaluate our method on 3D room layout estimation. Next, we evaluate the effectiveness of transferring the knowledge to 2D room layout estimation. For 3D layout estimation, we use metrics for depth map evaluation, including root of the mean squared error (rms), mean relative error (rel), Mean log10 error (log10), and the percentage of pixels with the ratio between the prediction and the ground truth smaller than $1.25$, $1.25^2$, and $1.25^3$. We also calculate a 3D corner error ($\rm e_{3D\_cor.}$), which represents the Euclidean distance between the 3D layout corners and ground truth in the camera coordinate system. The 3D coordinates can be calculated using the intrinsic parameters provided in the dataset. For 2D layout estimation, we use two standard metrics adopted by many benchmarks, including the pixel-wise segmentation error ($\rm e_{pix.}$) and the corner location error ($\rm e_{cor.}$)~\cite{yinda2016lsun}.

\subsection{Implementation Details}

The input images are resized to $228\times304$ using bilinear interpolation and the output size is $114\times152\times4$. The training images are augmented by random cropping and color jittering. The model is implemented using PyTorch~\cite{paszke2017automatic} with batch size of 32. We use Adam optimizer with an initial learning rate of $10^{-4}$ and a weight decay of $10^{-4}$. The network is trained for 200 epochs and the learning rate is reduced to 10\% for every 50 epochs. The values of the margins are set as $\delta_v=0.1$, $\delta_d=1.0$. The scale factor is set as $k=20$. The weights in the final loss functions are set as $\alpha=0.5$, $\beta=1$, $\eta=10$, $\theta=0.03$.

\subsection{Results on Matterport3D-Layout dataset.}

\noindent\textbf{3D layout performance.} The performance on the Matterport3D-Layout testing set is shown in Table~\ref{table:3d}. The existing layout estimation methods are mostly 2D based methods and cannot predict the 3D layout estimates. We compare to PlaneNet~\cite{liu2018planenet}, which is the state-of-the-art method for 3D planar reconstruction. The major difference between our method and PlaneNet is that PlaneNet directly estimates a 2D segmentation with fixed number of regions together with the instance-level 3D planar parameters, while we estimate the pixel-level surface parameters first and infer segmentation geometrically. The results in Table~\ref{table:3d} show that our method (GeoLayout-Ours) consistently outperforms PlaneNet on all the metrics. The reason might be that our method does not need to predict the error-prone 2D segmentation masks. In addition, the averaged instance-level surface parameters in GeoLayout are more robust against noise.

\begin{table*}
\begin{center}
\caption{Layout estimation results on the Matterport3D-Layout dataset.}
\label{table:3d}
\scalebox{1}[1]{
\begin{tabular}{c|ccccccccc}
\hline
Method & $\rm e_{pix.}$  & $\rm e_{cor.}$ & $\rm e_{3D\_cor.}$  & rms & rel &log10 &$\delta<1.25$ &$\delta<1.25^2$ &$\delta<1.25^3$ \\
\hline
PlaneNet~\cite{liu2018planenet}&6.89&5.29&14.00&0.520&0.134&0.057&0.846&0.954&0.984 \\
\makecell[c]{GeoLayout-Plane}&5.84&4.71&\textbf{12.05}&\textbf{0.448}&\textbf{0.109}&\textbf{0.046}&0.891&0.973&0.993 \\
GeoLayout-Ours &\textbf{5.24}&\textbf{4.36}&12.82&0.456&0.111&0.047&\textbf{0.892}&\textbf{0.975}&\textbf{0.994} \\
\hline
\end{tabular}
}
\end{center}
\end{table*}

We also compare to a version of our method using plane parameterization (GeoLayout-Plane). Instead of the proposed surface representation, we estimate the 4 parameters of typical planar equation (i.e. 3 for surface normal and 1 for the offset to the origin). In the testing stage, the predicted plane parameters are converted to the surface parameters using intrinsic parameters and the same layout generation process is performed to produce the layout estimates. GeoLayout-Plane shows comparable performance with GeoLayout-Ours. This indicates the network can successfully estimate surface parameters that already implicitly include camera intrinsics. However, as GeoLayout-Ours does not require the camera intrinsic parameters, it is more flexible in practice and can be easily run on images in the wild while GeoLayout-Plane cannot.

\begin{figure*}
\begin{center}
\includegraphics[width = 1.0\textwidth]{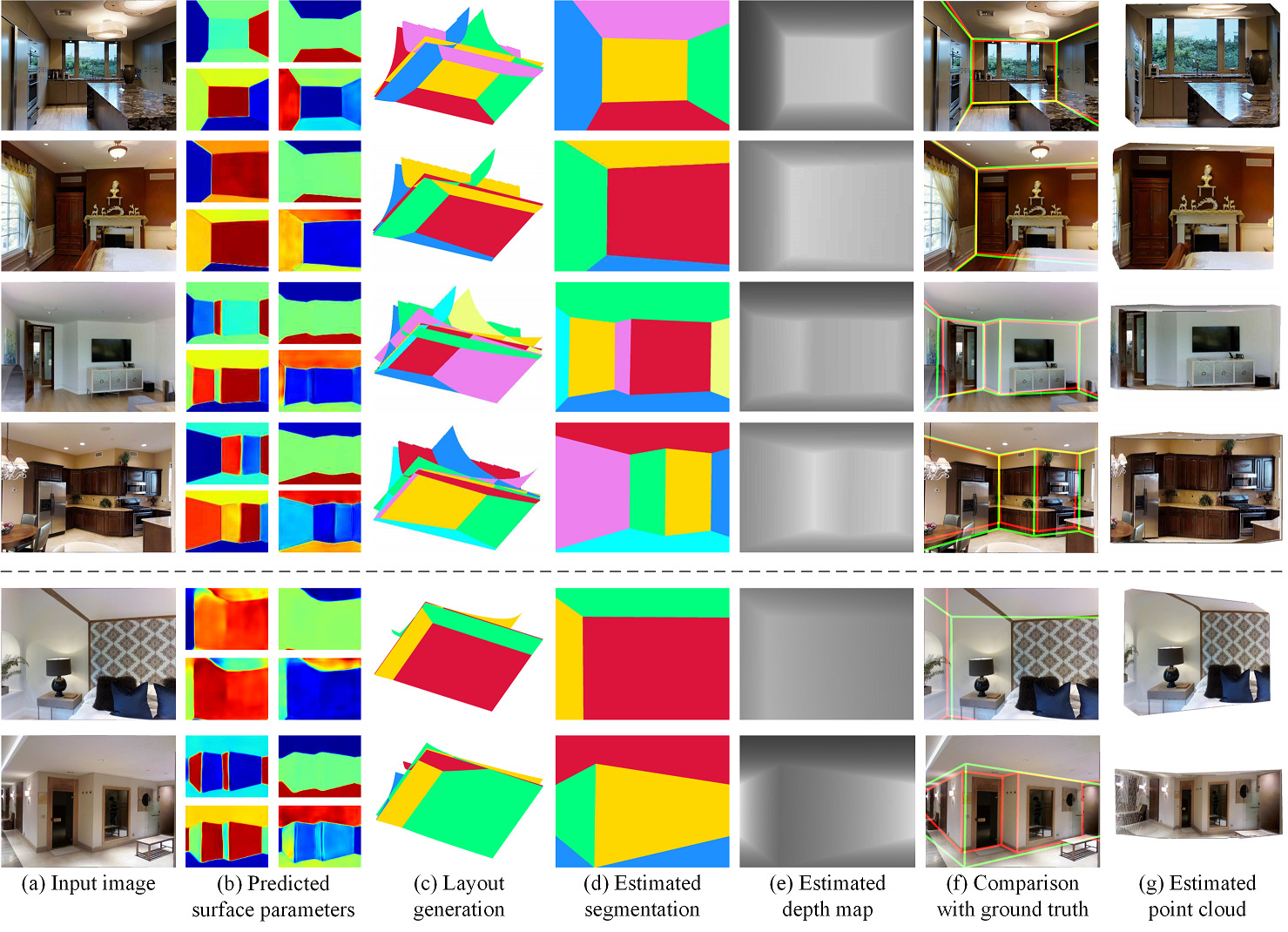}
\end{center}
\caption{Qualitative results on the Matterport3D-Layout dataset. The first two rows are cuboid rooms and the following two rows are non-cuboid rooms. Failure cases are shown in the last two rows.}
\label{fig:3dres}
\end{figure*}

\noindent\textbf{Qualitative results.} The qualitative results are given in Fig.~\ref{fig:3dres}. The predicted pixel-level surface parameters are shown in (b), with $p$ and $q$ shown in the first row, $r$ and $s$ shown in the second row. Based on the surface parameters, the depth maps of the surfaces are calculated and displayed in the $(u, v, Z)$ coordinate system as shown in (c). The estimated 2D segmentation and depth map are shown in (d) and (e), respectively. The comparison of the layout estimates (outlined by green) and the ground truth results (outlined by red) are shown in (f). We convert the estimated depth map into point cloud to better visualize the 3D layout estimates, as shown in (g). The first two rows are cuboid rooms and the following two rows are non-cuboid rooms. The results show that our method can reliably estimate the surface parameters and produces high quality layout estimates. Note that our method can handle the non-cuboid rooms with arbitrary number of walls. Two typical failure cases are shown in the last two rows. We found that most of the failure cases are either caused by the large prediction error of the surface parameters, or due to the error during clustering, especially for the non-cuboid rooms and those with more planar surfaces.

\subsection{Results on 2D layout datasets.}

\begin{figure*}
\begin{center}
\includegraphics[width = 1.0\textwidth]{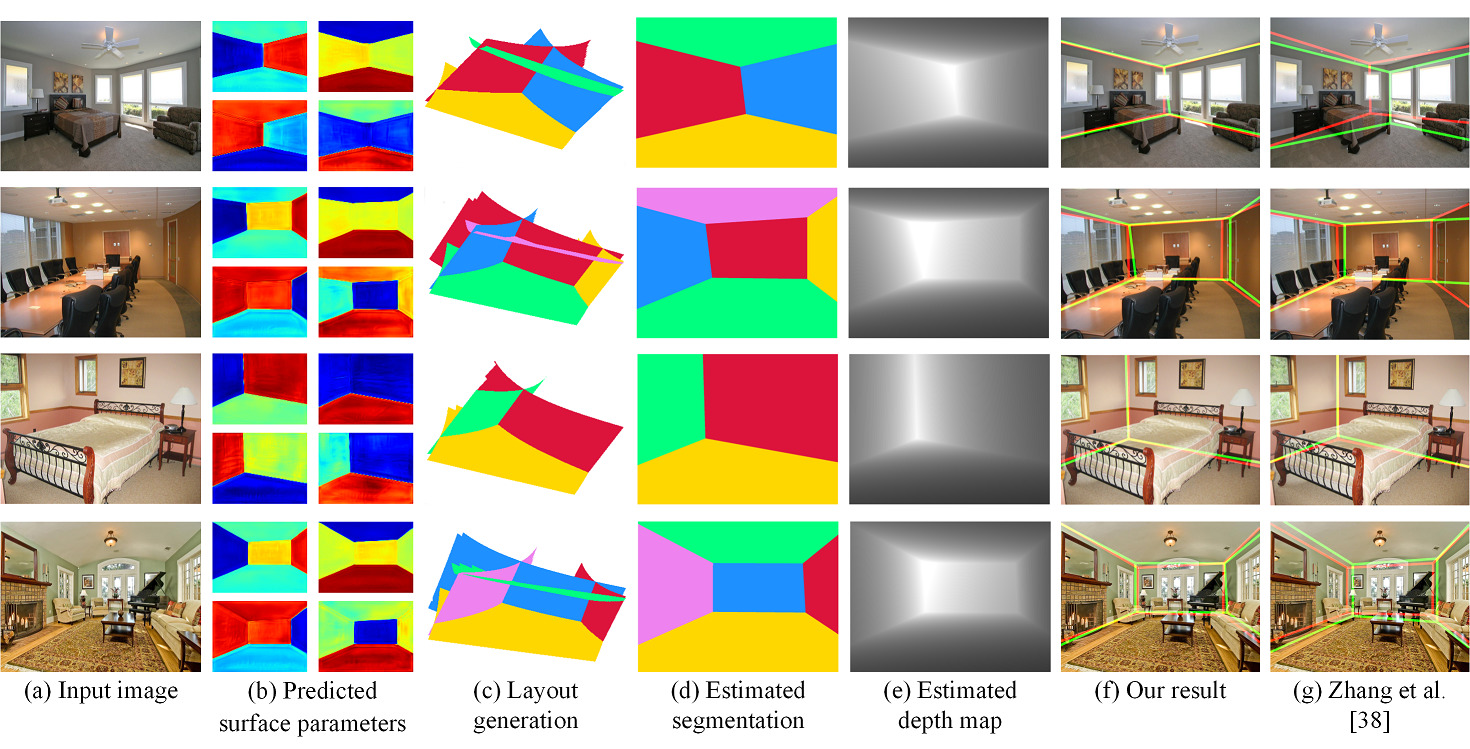}
\end{center}
\caption{Qualitative results on the LSUN validation set.}
\label{fig:2dres}
\end{figure*}

\noindent\textbf{Generalization to 2D Layout Estimation.} We verify our method on traditional 2D layout estimation benchmarks including Hedau~\cite{hedau2009recovering} and LSUN~\cite{yinda2016lsun}. We first directly run our model trained from Matterport3D-Layout dataset on the LSUN validation set without fine-tuning. The result is shown in Table~\ref{table:lsun_val} (w/o Fine-tune). The model still produces reasonable results, which indicates some generalization capability. We then fine-tune our model on LSUN as described in Section~\ref{sec:2d}, and the performance is significantly improved (w/ Fine-tune). This indicates that the model can be effectively trained on 2D layout dataset with the proposed learning strategy.

\begin{table}
\begin{center}
\caption{Comparison of the model w/wo fine-tuning on LSUN validation dataset.}
\label{table:lsun_val}
\setlength{\tabcolsep}{2mm}{
\begin{tabular}{lcc}
\hline
Setting& $\rm e_{pix.}$ (\%) & $\rm e_{cor.}$ (\%)  \\
\hline
w/o Fine-tune &12.67&8.12 \\
w/ Fine-tune &\textbf{6.10}&\textbf{4.66} \\
\hline
\end{tabular}}
\end{center}
\end{table}

\noindent\textbf{2D layout performance.} We compare our fine-tuned model on LSUN test set and Hedau dataset to other state-of-the-art methods in Table~\ref{table:hedau}. The LSUN performance is reported by the dataset owner on withheld ground truth to prevent over-fitting. Our method achieves the best performance on LSUN dataset and the second best performance on Hedau dataset. Such result shows that incorporating 3D knowledge and geometric reasoning to layout estimation is beneficial and can significantly improve the 2D layout estimation performance.

\begin{table}
\begin{center}
\caption{Layout estimation performance on LSUN~\cite{yinda2016lsun} and  Hedau~\cite{hedau2009recovering} datasets.}
\label{table:hedau}
\setlength{\tabcolsep}{1mm}{
\begin{tabular}{lccc}
\hline
Method & LSUN~$\rm e_{pix.}$ (\%) & LSUN~$\rm e_{cor.}$ (\%) & Hedau~$\rm e_{pix.}$  (\%)  \\
\hline
Hedau \emph{et~al.} (2009)~\cite{hedau2009recovering}&24.23&15.48&21.20 \\
Mallya \emph{et~al.} (2015)~\cite{mallya2015learning}&16.71&11.02&12.83 \\
Dasgupta \emph{et~al.} (2016)~\cite{dasgupta2016delay}&10.63&8.20&9.73 \\
Ren \emph{et~al.} (2016)~\cite{ren2016coarse}&9.31&7.95&8.67 \\
Lee \emph{et~al.} (2017)~\cite{lee2017roomnet}&9.86&6.30&8.36 \\
Hirzer \emph{et~al.} (2020)~\cite{Hirzer_2020_WACV}&7.79&5.84&7.44 \\
Kruzhilov \emph{et~al.} (2019)~\cite{kruzhilov2019double}&6.72&5.11&7.85 \\
Zhang \emph{et~al.} (2019)~\cite{zhang2019edge}&6.58&5.17&7.36 \\
Hsiao \emph{et~al.} (2019)~\cite{hsiao2019flat2layout}&6.68&4.92&\textbf{5.01} \\
GeoLayout&\textbf{6.09}&\textbf{4.61}&7.16 \\
\hline
\end{tabular}}
\end{center}
\end{table}

\noindent\textbf{Qualitative results.} Fig.~\ref{fig:2dres} shows the visual results on the LSUN validation set, with the results of Zhang et
al.~\cite{zhang2019edge} for comparison. As can be seen, our method is less error-prone and generally produces more precise results than~\cite{zhang2019edge}.

\section{Conclusion}

This paper proposed a novel geometry driven method for indoor layout estimation. The key idea is to learn the depth maps of planar surfaces and then generate the layout by applying geometric rules. We demonstrated that the model could be trained effectively using either 2D or 3D ground truths. The proposed method achieved state-of-the-art performance on benchmark datasets for both 2D and 3D layout. We also presented a new dataset with 3D layout ground truth, which we believe is beneficial to the field of room layout estimation.~\\

\noindent\textbf{Acknowledgements} This work was supported in part by the National Natural Science Foundation of China under Grant 61991411, and Grant U1913204, in part by the National Key Research and Development Plan of China under Grant 2017YFB1300205, and in part by the Shandong Major Scientific and Technological Innovation Project (MSTIP) under Grant 2018CXGC1503. We thank the LSUN organizer for the benchmarking service.

%

\clearpage
%
%

\begin{thebibliography}{10}
\providecommand{\url}[1]{\texttt{#1}}
\providecommand{\urlprefix}{URL }
\providecommand{\doi}[1]{https://doi.org/#1}

\bibitem{barinova2008fast}
Barinova, O., Konushin, V., Yakubenko, A., Lee, K., Lim, H., Konushin, A.: Fast
  automatic single-view 3-d reconstruction of urban scenes. In: European
  Conference on Computer Vision. pp. 100--113. Springer (2008)

\bibitem{camplani2013depth}
Camplani, M., Mantecon, T., Salgado, L.: Depth-color fusion strategy for 3-d
  scene modeling with kinect. IEEE Transactions on Cybernetics  \textbf{43}(6),
   1560--1571 (2013)

\bibitem{Matterport3D}
Chang, A., Dai, A., Funkhouser, T., Halber, M., Niessner, M., Savva, M., Song,
  S., Zeng, A., Zhang, Y.: Matterport3d: Learning from rgb-d data in indoor
  environments. International Conference on 3D Vision (3DV)  (2017)

\bibitem{choi2013understanding}
Choi, W., Chao, Y.W., Pantofaru, C., Savarese, S.: Understanding indoor scenes
  using 3d geometric phrases. In: Proceedings of the IEEE Conference on
  Computer Vision and Pattern Recognition. pp. 33--40 (2013)

\bibitem{dasgupta2016delay}
Dasgupta, S., Fang, K., Chen, K., Savarese, S.: Delay: Robust spatial layout
  estimation for cluttered indoor scenes. In: Proceedings of the IEEE
  Conference on Computer Vision and Pattern Recognition. pp. 616--624 (2016)

\bibitem{de2017semantic}
De~Brabandere, B., Neven, D., Van~Gool, L.: Semantic instance segmentation with
  a discriminative loss function. arXiv preprint arXiv:1708.02551  (2017)

\bibitem{del2013understanding}
Del~Pero, L., Bowdish, J., Kermgard, B., Hartley, E., Barnard, K.:
  Understanding bayesian rooms using composite 3d object models. In:
  Proceedings of the IEEE Conference on Computer Vision and Pattern
  Recognition. pp. 153--160 (2013)

\bibitem{delage2007automatic}
Delage, E., Lee, H., Ng, A.Y.: Automatic single-image 3d reconstructions of
  indoor manhattan world scenes. In: Robotics Research, pp. 305--321. Springer
  (2007)

\bibitem{fischler1981random}
Fischler, M.A., Bolles, R.C.: Random sample consensus: a paradigm for model
  fitting with applications to image analysis and automated cartography.
  Communications of the ACM  \textbf{24}(6),  381--395 (1981)

\bibitem{fouhey2014unfolding}
Fouhey, D.F., Gupta, A., Hebert, M.: Unfolding an indoor origami world. In:
  European Conference on Computer Vision. pp. 687--702. Springer (2014)

\bibitem{guo2015predicting}
Guo, R., Zou, C., Hoiem, D.: Predicting complete 3d models of indoor scenes.
  arXiv preprint arXiv:1504.02437  (2015)

\bibitem{haines2014recognising}
Haines, O., Calway, A.: Recognising planes in a single image. IEEE transactions
  on pattern analysis and machine intelligence  \textbf{37}(9),  1849--1861
  (2014)

\bibitem{hedau2009recovering}
Hedau, V., Hoiem, D., Forsyth, D.: Recovering the spatial layout of cluttered
  rooms. In: Proceedings of the IEEE International Conference on Computer
  Vision. pp. 1849--1856. IEEE (2009)

\bibitem{Hirzer_2020_WACV}
Hirzer, M., Lepetit, V., ROTH, P.: Smart hypothesis generation for efficient
  and robust room layout estimation. In: The IEEE Winter Conference on
  Applications of Computer Vision (WACV) (March 2020)

\bibitem{hsiao2019flat2layout}
Hsiao, C.W., Sun, C., Sun, M., Chen, H.T.: Flat2layout: Flat representation for
  estimating layout of general room types. arXiv preprint arXiv:1905.12571
  (2019)

\bibitem{hu2019revisiting}
Hu, J., Ozay, M., Zhang, Y., Okatani, T.: Revisiting single image depth
  estimation: Toward higher resolution maps with accurate object boundaries.
  In: 2019 IEEE Winter Conference on Applications of Computer Vision (WACV).
  pp. 1043--1051. IEEE (2019)

\bibitem{izadinia2017im2cad}
Izadinia, H., Shan, Q., Seitz, S.M.: Im2cad. In: Proceedings of the IEEE
  Conference on Computer Vision and Pattern Recognition. pp. 5134--5143 (2017)

\bibitem{karsch2011rendering}
Karsch, K., Hedau, V., Forsyth, D., Hoiem, D.: Rendering synthetic objects into
  legacy photographs. In: ACM Transactions on Graphics (TOG). vol.~30, p.~157.
  ACM (2011)

\bibitem{kruzhilov2019double}
Kruzhilov, I., Romanov, M., Babichev, D., Konushin, A.: Double refinement
  network for room layout estimation. In: Asian Conference on Pattern
  Recognition. pp. 557--568. Springer (2019)

\bibitem{lee2017roomnet}
Lee, C.Y., Badrinarayanan, V., Malisiewicz, T., Rabinovich, A.: Roomnet:
  End-to-end room layout estimation. In: Proceedings of the IEEE International
  Conference on Computer Vision. pp. 4865--4874 (2017)

\bibitem{lee2009geometric}
Lee, D.C., Hebert, M., Kanade, T.: Geometric reasoning for single image
  structure recovery. In: Proceedings of the IEEE Conference on Computer Vision
  and Pattern Recognition. pp. 2136--2143. IEEE (2009)

\bibitem{liu2019planercnn}
Liu, C., Kim, K., Gu, J., Furukawa, Y., Kautz, J.: Planercnn: 3d plane
  detection and reconstruction from a single image. In: Proceedings of the IEEE
  Conference on Computer Vision and Pattern Recognition. pp. 4450--4459 (2019)

\bibitem{liu2018planenet}
Liu, C., Yang, J., Ceylan, D., Yumer, E., Furukawa, Y.: Planenet: Piece-wise
  planar reconstruction from a single rgb image. In: Proceedings of the IEEE
  Conference on Computer Vision and Pattern Recognition. pp. 2579--2588 (2018)

\bibitem{liu2015rent3d}
Liu, C., Schwing, A.G., Kundu, K., Urtasun, R., Fidler, S.: Rent3d: Floor-plan
  priors for monocular layout estimation. In: Proceedings of the IEEE
  Conference on Computer Vision and Pattern Recognition. pp. 3413--3421. IEEE
  (2015)

\bibitem{mallya2015learning}
Mallya, A., Lazebnik, S.: Learning informative edge maps for indoor scene
  layout prediction. In: Proceedings of the IEEE International Conference on
  Computer Vision. pp. 936--944 (2015)

\bibitem{martin20143d}
Martin-Brualla, R., He, Y., Russell, B.C., Seitz, S.M.: The 3d jigsaw puzzle:
  Mapping large indoor spaces. In: European Conference on Computer Vision. pp.
  1--16. Springer (2014)

\bibitem{micusk2008towards}
Micusk, B., Wildenauer, H., Vincze, M.: Towards detection of orthogonal planes
  in monocular images of indoor environments. In: 2008 IEEE International
  Conference on Robotics and Automation. pp. 999--1004. IEEE (2008)

\bibitem{paszke2017automatic}
Paszke, A., Gross, S., Chintala, S., Chanan, G., Yang, E., DeVito, Z., Lin, Z.,
  Desmaison, A., Antiga, L., Lerer, A.: Automatic differentiation in pytorch
  (2017)

\bibitem{pero2012bayesian}
Pero, L.D., Bowdish, J., Fried, D., Kermgard, B., Hartley, E., Barnard, K.:
  Bayesian geometric modeling of indoor scenes. In: Proceedings of the IEEE
  Conference on Computer Vision and Pattern Recognition. pp. 2719--2726. IEEE
  (2012)

\bibitem{ramalingam2013manhattan}
Ramalingam, S., Pillai, J., Jain, A., Taguchi, Y.: Manhattan junction catalogue
  for spatial reasoning of indoor scenes. In: Proceedings of the IEEE
  Conference on Computer Vision and Pattern Recognition. pp. 3065--3072 (2013)

\bibitem{ren2016coarse}
Ren, Y., Li, S., Chen, C., Kuo, C.C.J.: A coarse-to-fine indoor layout
  estimation (cfile) method. In: Asian Conference on Computer Vision. pp.
  36--51. Springer (2016)

\bibitem{russell2008labelme}
Russell, B.C., Torralba, A., Murphy, K.P., Freeman, W.T.: Labelme: a database
  and web-based tool for image annotation. International journal of computer
  vision  \textbf{77}(1-3),  157--173 (2008)

\bibitem{long2015fully1}
Shelhamer, E., Long, J., Darrell, T.: Fully convolutional networks for semantic
  segmentation. IEEE Transactions on Pattern Analysis and Machine Intelligence
  \textbf{39}(4),  640--651 (2017)

\bibitem{wang2013discriminative}
Wang, H., Gould, S., Roller, D.: Discriminative learning with latent variables
  for cluttered indoor scene understanding. Communications of the ACM
  \textbf{56}(4),  92--99 (2013)

\bibitem{xiao2014reconstructing}
Xiao, J., Furukawa, Y.: Reconstructing the world's museums. International
  Journal of Computer Vision  \textbf{110}(3),  243--258 (2014)

\bibitem{yang2018recovering}
Yang, F., Zhou, Z.: Recovering 3d planes from a single image via convolutional
  neural networks. In: Proceedings of the European Conference on Computer
  Vision (ECCV). pp. 85--100 (2018)

\bibitem{yu2019single}
Yu, Z., Zheng, J., Lian, D., Zhou, Z., Gao, S.: Single-image piece-wise planar
  3d reconstruction via associative embedding. arXiv preprint arXiv:1902.09777
  (2019)

\bibitem{zhang2019edge}
Zhang, W., Zhang, W., Gu, J.: Edge-semantic learning strategy for layout
  estimation in indoor environment. IEEE transactions on cybernetics  (2019)

\bibitem{yinda2016lsun}
Zhang, Y., Yu, F., Song, S., Xu, P., Seff, A., Xiao, J.: Largescale scene
  understanding challenge: Room layout estimation
  \url{http://lsun.cs.princeton.edu/2016/}

\bibitem{zhao2017physics}
Zhao, H., Lu, M., Yao, A., Guo, Y., Chen, Y., Zhang, L.: Physics inspired
  optimization on semantic transfer features: An alternative method for room
  layout estimation. arXiv preprint arXiv:1707.00383  (2017)

\bibitem{zhao2017pyramid}
Zhao, H., Shi, J., Qi, X., Wang, X., Jia, J.: Pyramid scene parsing network.
  In: Proceedings of the IEEE conference on computer vision and pattern
  recognition. pp. 2881--2890 (2017)

\bibitem{zhao2013scene}
Zhao, Y., Zhu, S.C.: Scene parsing by integrating function, geometry and
  appearance models. In: Proceedings of the IEEE Conference on Computer Vision
  and Pattern Recognition. pp. 3119--3126 (2013)

\bibitem{zou2018layoutnet}
Zou, C., Colburn, A., Shan, Q., Hoiem, D.: Layoutnet: Reconstructing the 3d
  room layout from a single rgb image. In: Proceedings of the IEEE Conference
  on Computer Vision and Pattern Recognition. pp. 2051--2059 (2018)

\end{thebibliography}

\nocite{*}

\end{document}